\documentclass[10pt,letterpaper]{article}
\usepackage{cogsci}
\usepackage{graphicx}
\usepackage{hyperref}
\usepackage{xcolor}
\usepackage{tikz}
\usepackage{dblfloatfix}
\usetikzlibrary{patterns}
\usepackage{pgfplots, pgfplotstable}
\graphicspath{{Images/}}
\newcounter{groupcount}
\pgfplotsset{
    draw group line/.style n args={5}{
        after end axis/.append code={
            \setcounter{groupcount}{0}
            \pgfplotstableforeachcolumnelement{#1}\of\datatable\as\cell{%
                \def\temp{#2}
                \ifx\temp\cell
                    \ifnum\thegroupcount=0
                        \stepcounter{groupcount}
                        \pgfplotstablegetelem{\pgfplotstablerow}{X}\of\datatable
                        \coordinate [yshift=#4] (startgroup) at (axis cs:\pgfplotsretval,0);
                    \else
                        \pgfplotstablegetelem{\pgfplotstablerow}{X}\of\datatable
                        \coordinate [yshift=#4] (endgroup) at (axis cs:\pgfplotsretval,0);
                    \fi
                \else
                    \ifnum\thegroupcount=1
                        \setcounter{groupcount}{0}
                        \draw [
                            shorten >=-#5,
                            shorten <=-#5
                        ] (startgroup) -- node [anchor=base, yshift=0.5ex] {#3} (endgroup);
                    \fi
                \fi
            }
            \ifnum\thegroupcount=1
                        \setcounter{groupcount}{0}
                        \draw [
                            shorten >=-#5,
                            shorten <=-#5
                        ] (startgroup) -- node [anchor=base, yshift=0.5ex] {#3} (endgroup);
            \fi
        }
    }
}

\cogscifinalcopy 

\usepackage{pslatex}
\usepackage{apacite}
\usepackage{float} 

\title{A Grounded Approach to Modeling Generic Knowledge Acquisition}

\author{{\large \bf Deniz Beser (beser@isi.edu)} \\
        {\large \bf Joe Cecil (cecil@isi.edu)} \\
        {\large \bf Marjorie Freedman (mrf@isi.edu)} \\
        {\large \bf Jacob Lichtefeld (jacobl@isi.edu)} \\
          Information Sciences Institute, University of Southern California \\
          Marina del Rey, CA 90292, USA
    \AND 
        {\large \bf Mitch Marcus (mitch@cis.upenn.edu)} \\
      Department of Computer and Information Science, University of Pennsylvania \\
      Philadelphia, PA 19104, USA
  \AND 
      {\large \bf Sarah Payne (paynesa@sas.upenn.edu)} \\ {\large \bf Charles Yang (charles.yang@ling.upenn.edu)} \\
      Departments of Linguistics and Computer and Information Science,
      University of Pennsylvania \\
        Philadelphia, PA 19104, USA
  }

\begin{document}

\maketitle

\begin{abstract}
We introduce and implement a cognitively plausible model for learning from generic language, statements that express generalizations about members of a category and are an important aspect of concept development in language acquisition \cite{carlson1995generic, gelman2009learning}. We extend a computational framework designed to model grounded language acquisition by introducing the \textit{concept network}. This new layer of abstraction enables the system to encode knowledge learned from generic statements and represent the associations between concepts learned by the system. Through three tasks that utilize the concept network, we demonstrate that our extensions to ADAM can acquire generic information and provide an example of how ADAM can be used to model language acquisition.
\footnote{Our code is available at 
\href{https://github.com/isi-vista/adam}{https://github.com/isi-vista/adam}}

\end{abstract}

\section{Introduction}

Generics, statements in the form of \textit{``bananas are yellow"} or \textit{``birds fly"}, express generalizations about members of a category, and are frequent in everyday language \cite{carlson1977reference, carlson1995generic}. These expressions refer to a category as a whole (e.g\textit{ birds}), as opposed to a single instance of a category (e.g \textit{a bird}), and while referring to a conceptual category as a whole, generic statements may assert information that while typical, does not cover all instances, and hence are not necessarily invalidated by counter-examples \cite{mccawley1981everything, gelman2004learning}. For example, the statement \textit{``birds fly"} is not invalidated by the existence of penguins, a bird that cannot fly. Moreover, experience with only a single instance of a conceptual category can be sufficient to acquire generic knowledge \cite{carlson1995generic, prasada2000acquiring}. Besides the prevalence and the expressive power of generics in language, generics also play an important role in child language acquisition. During conceptual development, generic statements complement observational learning and help construct conceptual knowledge, allowing children to learn hierarchical information that cannot be learned from the world with experience alone \cite{cimpian2009information, gelman2009learning}.

\citeA{prasada2000acquiring} emphasized the acquisition problem posed by the imprecise nature of generic language and our ability to learn generic knowledge from minimal experience, outlining that the requirements of a formal system for acquiring generic knowledge are to complement other learning mechanisms and to establish relationships between categories and properties. By age 2, children manifest the capabilities of such a system, as they can recognize category labels \cite{graham2004thirteen} and make use of names to classify objects, even if an instance is atypical for a given category \cite{nazzi2001linguistic, jaswal2007looks}. Considering how generic utterances such as \textit{``penguins are birds"} can override interpretation of perceptual features and allow learning hierarchical knowledge even from a young age \cite{gelman2009learning, Cimpian2010}, it is clear that language can inform perceptual learning, making generics a critical component of language acquisition worth modeling.


Inspired by the plethora of research on generics and child language development, we introduce and implement a developmentally plausible model for learning concepts from generic language that assumes no prior conceptual knowledge, learns meanings of words and concepts in a manner similar to infants (from the ground up), and demonstrates the capacity to make inference with the learned concepts. The importance of generic language has been recognized in the artificial intelligence and natural language processing community for tasks that involve knowledge acquisition, ontology development, and semantic inference \cite<e.g.,>{Reiter2010, Friedrich2015, Sedghi2018}. These approaches generally make use of large-scale resources and employ methods such as supervised learning that are not suitable for modeling child language and conceptual development. We base our system on ADAM, a software platform for modeling early language acquisition \cite{adam2021}. Through pairs of perceptual representations of situations and linguistic utterances describing these situations, ADAM learns interpretable representations for \textit{concepts} such as objects, attributes, relations, and actions (see Model and Representation section for further detail). ADAM's cognitively plausible design choices regarding perceptual representations \cite{marr1982vision, biederman1987recognition} and simple word/pattern learning mechanisms \cite{webster-marcus-1989-automatic, Yu2012, stevens2017pursuit} lay the groundwork for modeling higher level semantics in child language acquisition. Given the interplay between generic statements, concepts, and observational learning in natural language \cite{carlson1977reference, carlson1995generic, gelman2009learning} and ADAM's power to learn meanings of concepts and identify them across situations, we consider ADAM to be a promising system for modeling acquisition of generic knowledge and the resultant semantic category inference.

We expand ADAM's modeling capabilities to capture semantic associations and generics by introducing a generic learner module and combining ADAM's representations with an additional layer of abstraction, a network data structure called \textit{the concept network}. The concept network organizes the associations between concepts and learns hierarchies and properties about the concepts through observation as well as generic utterances. Through three tasks that use generic language across different learning curricula, we demonstrate that ADAM, coupled with the generic learner module and the concept network, can acquire generic knowledge, establish semantic certainty with generics language, and make category inference. Our demonstrations provide an example of how ADAM can be used to model language acquisition.

\section{Model and Representation}
\subsection{Meaning Representation in ADAM}
\begin{figure}
    \centering
    \includegraphics[width=8cm]{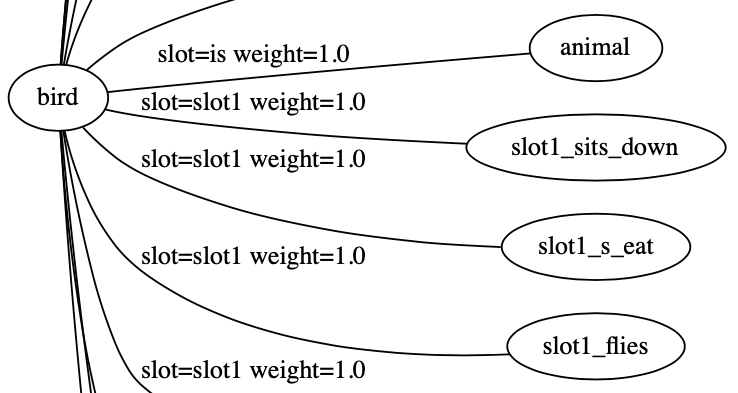}
    \hrule
    \centering
    \includegraphics[width=8.5cm]{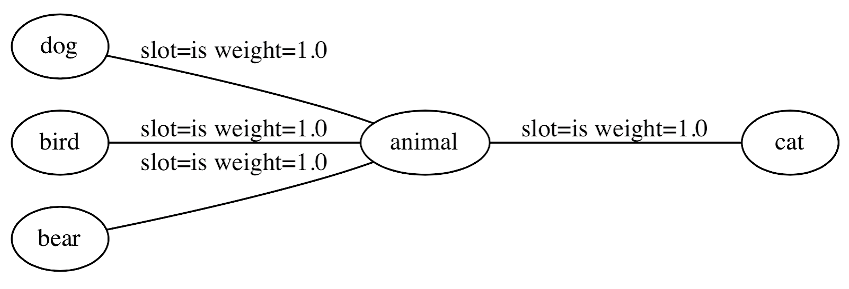}
    \caption{Examples of a concept network. Nodes represent concepts (e.g. \textit{bird}, \textit{animal}, \textit{sitting down}). The edges are labeled by the semantic relation between the concepts (slot), and the association strength (weight). The top shows a portion of the concept graph for \textit{bird}. The bottom shows \textit{animal} and its neighbors.}
    \label{fig:cn-graph}
\end{figure}

ADAM is a software platform for experiments in child language learning. The system can process a range of expressions covering a very young learner's vocabulary and grammar, such as objects (\textit{``a ball"}), adjectives (\textit{``a red ball''}), prepositions (\textit{``a ball on a table"}) and actions (\textit{``Mom rolls a ball"}). ADAM uses \textit{perception graphs} to represent situations (e.g. a cup sitting on a table) that it perceives. Perception graphs consist of perceptually plausible components, such as geons \cite{biederman1987recognition}, body parts, colors, and regions. For example, when the model observes a cup on the table, the model perceives a structured graph that consists of the cup's color, hollow shape, its handle, and its position relative to the table. The model learns patterns over observed perception graphs. The patterns represent hypotheses about the meaning of individual \textit{concepts} and are consolidated throughout the learning process. For example, the learner could perceive perception graphs and utterances from multiple situations that involve a bear, such as hearing \textit{ ``a brown bear"} while observing a bear by itself, hearing \textit{``a bear sits"} while observing a bear sitting, and hearing \textit{``two bears"} while observing two bears, and eventually learn a representation of \textit{``bear"}'s meaning. ADAM uses the learned mapping between linguistic structures and meaning representations to describe new scenes.  Before learning, we define a configurable curriculum that pairs observations and descriptions. During testing, the perception input is generated without the descriptions.

\subsection{The Concept Network}
We extend ADAM's representations of concepts and patterns with the concept network, a graph-based data structure that represents learned concepts (e.g, objects, attributes, and actions) as nodes and the semantic associations between them as edges. The concept network enables one to see whether any two concepts are related, what their relation is, and how strong this relation is, by looking at the edge connecting the two concepts. Contrary to ADAM's perception graphs and patterns that are used to describe the components of a scene or perceptual properties of a perceived object, the concept network represents the overall semantic and conceptual knowledge of the learner learned over time. We represent each concept as a single node in the network; for example, \textit{bird} is represented by a single concept node in the concept network, and other semantically related concepts are its neighbors. Figure \ref{fig:cn-graph} (top) visualizes a section of the learned concept network structure for the \textit{bird} concept. Using the concept network, we can infer that a concept such as \textit{bird} is an \textit{animal}, and that it is associated with the concept \textit{fly}. Since these associations are formed through the learner's observations of occurrences of concepts throughout learning, the association between the \textit{bird} and \textit{fly} concepts would suggest that a bird was observed while flying. Similarly, the network can be used to observe the categories formed by the learner; Figure \ref{fig:cn-graph} (bottom) shows the \textit{animal} node and its neighbors. To construct the nodes and edges in this network, while perceiving concepts and semantic relations between them as the learner observes situations as described in \citeA{adam2021}, the learner also simultaneously creates nodes for concepts (e.g \textit{bird}, \textit{fly}) and edges to associate concepts in the network. The concept network consists of about 250 concepts when generated with ADAM's standard curriculum (see \citeA{adam2021} for a complete list of content descriptors).


\subsubsection{Edges and Semantic Associations}
The edges in the concept network represent the semantic relation between the two concepts and the association strength between them. For action concepts, the relation often denotes an argument relation with the neighboring concept; the argument relation is labeled with \textit{slots} that describe a slot an argument can take in a phrase. For instance, given \textit{``Mom drinks juice"}, \textit{Mom} has argument slot position \textit{slot-1} with \textit{drink}, while \textit{juice} has \textit{slot-2}. The strengths in the network start from \textit{0} and have a maximum of \textit{1.0}. Throughout learning, the association strength between two concepts is updated with each co-occurrence of the two concepts, using the plateauing update function \(a = a + 0.2 * (1- a) \). The edges that represent semantic associations initialized with generic statements have maximum association strength, since we tend to believe what we are told is true. These edges are also marked with a slot label that matches the argument relation of the statement, or the \textit{is} label if the statement is a predicate.

\subsubsection{Matrix Representation}
We represent the network as a weighted adjacency matrix to facilitate rich operations such as vector similarity between concepts. In order to preserve argument relation information in the edges in the matrix, each action concept in the concept network is first translated to multiple concepts that include the argument relation (e.g \textit{drink slot 1} to represent drink in \textit{Mom drinks} and \textit{drink - slot 2} to represent drink in \textit{``drinking juice"}). 

\subsubsection{Concept Similarity} 
\begin{figure}
    \centering
    \includegraphics[width=8cm]{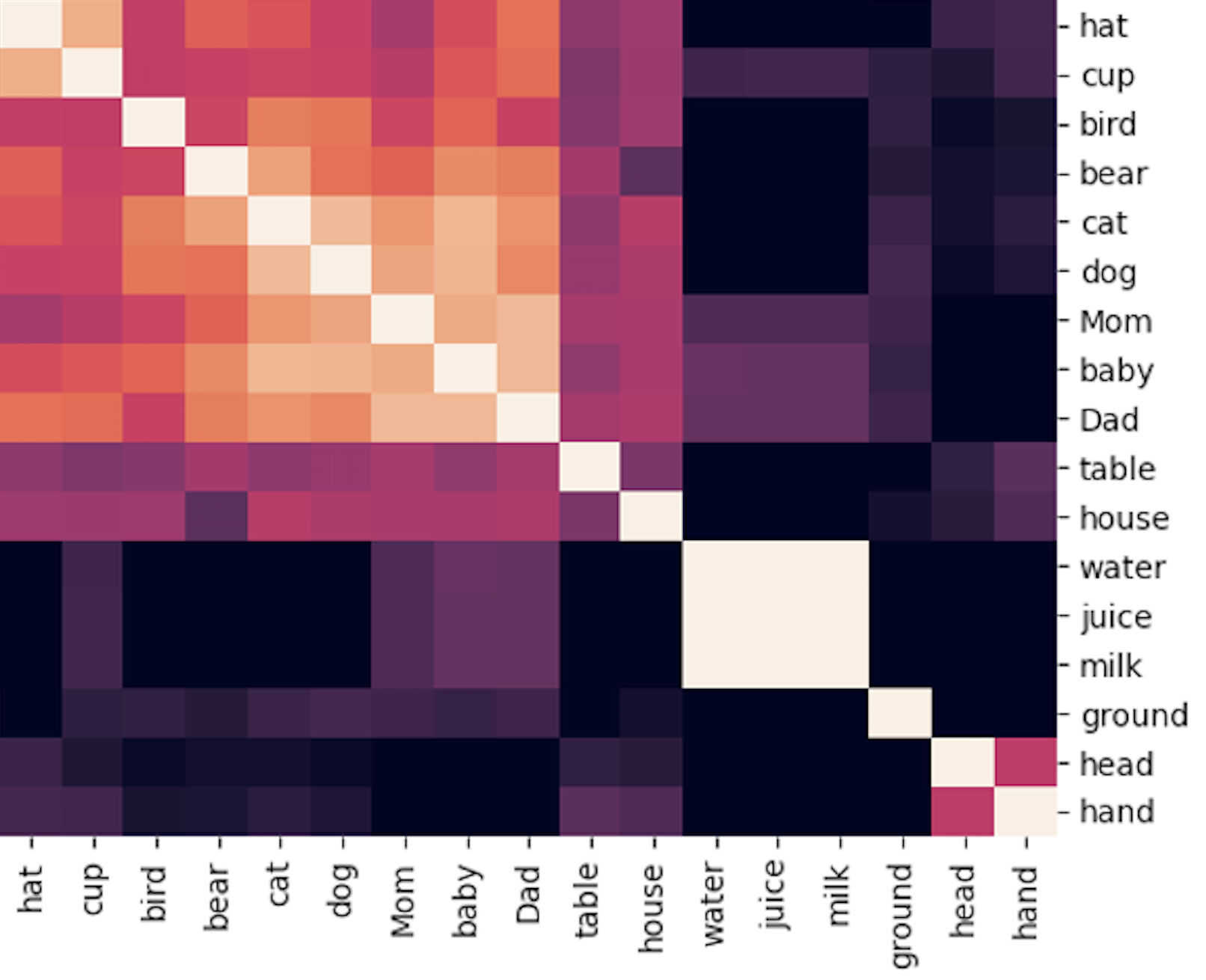}
    \caption{A qualitative demonstration of how hierarchical categories can be measured with concept similarity. We see the formation of categories in the concept space, e.g animate objects \textit{(cat, dog, Mom, baby, Dad)}, liquids \textit{(water, juice, milk)}, and body parts \textit{(head, hand)}. The vector representations of learned concepts were clustered with hierarchical clustering method \textit{clustermap} \protect\cite{mullner2011modern, waskom2020seaborn}.} 
    \label{fig:clustering}
\end{figure}

We use concept similarity for evaluation. Each concept is represented as an adjacency vector extracted from the weighted adjacency matrix. Vectors for \textit{category concepts} (e.g\textit{ animal}) are represented as the average of the vectors of all members of that category. Similarity between pairs of concepts is measured using cosine similarity. Measuring concept similarity provides insight into the the hierarchical structures that naturally emerge throughout learning, as visualized in Figure \ref{fig:clustering} which shows a clustered heatmap of the vector representations of concepts that are learned through a curriculum of objects and actions, without generics. We see the formation of categories that correspond to body parts, liquids, and animate objects.

 
\subsection{Generic Utterances}
To teach generic language and explicit categories to the learner, we present simple scenes paired with generic utterances such as \textit{``bears sit"} and \textit{``bears are brown"}. Since generic language encodes generic knowledge (e.g \textit{``birds fly"} stipulates that birds generally can fly), inputs to the learner that are in generic form maximize the corresponding association strength when observed by the learner. For instance, while the non-generic utterance \textit{``a bear sits"} increases the association strength between between \textit{bear} and \textit{sit}, observing \textit{``bears sit"} maximizes it.

\subsubsection{Generic Learner Module}
The ADAM system is built of learning modules. Each module targets a specific type of learning, such as objects, actions, and relations. We build a generic learner module to enable learning from generic utterances. The generic learner first verifies that the utterance is a generic by checking whether all the recognized nouns in the utterance are bare plurals, an indicative property of generics in English \cite{lyons1977semantics, gelman2004learning}. 
Once confirmed, the learner recognizes the concepts mentioned in the utterance, and forms a semantic connection between these concepts in the concept network. The association strength of this connection is maximal, which implies a semantic certainty that is learned from generic input.

In the special case where the generic statement is a predicate containing a previously unknown category, such as \textit{animal} in \textit{``dogs are animals,"} we create a new category concept node in the concept network and associate the object concept \textit{(dog)} with the category concept \textit{(animal)}. If a novel object concept appears in a generic statement as a member of a category concept, such as \textit{wug} in \textit{``wugs are animals,"} we create a new object concept and associate it with the category concept as well as the features of the members of the category. Overall, the module can interpret statements in the form of \textit{``birds fly,"} \textit{``birds are animals"} and \textit{``wugs are animals"}.

\section{Model Evaluation}
We present three tasks to illustrate the behavior of the system across different learning conditions.
\subsection{Task 1: Generic Color Predicates}
The generic color predicates task shows how the system can learn generics by establishing stronger associations between learned concepts. In other words, we measure how the generic input can change the learner's understanding of the world by influencing semantic connections. In this task, we first teach objects and arbitrary colors to the system with non-generic utterances (e.g \textit{a red truck}) based on the standard object and color curricula provided in the ADAM system. We then evaluate the associations between the object and color concepts that the system has learned. Then, we pick colors that are typical for the object concepts and teach them with generic statements (\textit{``watermelons are green," ``papers are white," ``cookies are light brown"}), and evaluate the associations between the object and color concepts again. In the non-generic training phase, each object appears with a variety of arbitrary colors (e.g cookies with blue, green, and light-brown colors), which we expect will lead the model to learn associations between an object and many colors. We expect the presentation of generic input to yield stronger associations between the object and the color used in a generic utterance.

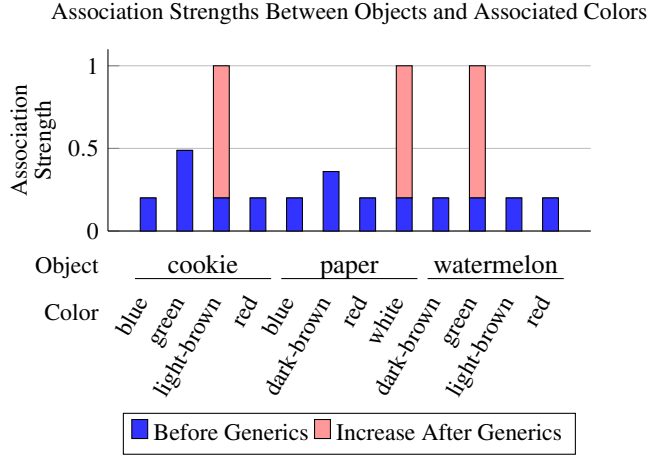
\begin{figure}[h]
\centering
\begin{tikzpicture}
\pgfplotstableread{
X   Object   Color Before After
1	cookie	 blue	0.2	0
2	cookie	 green	0.488	0
3	cookie	 light-brown	0.2	0.8
4	cookie	 red	0.2	0
5	paper	 blue	0.2	0
6	paper	 dark-brown	0.36	0
7	paper	 red	0.2	0
8	paper	 white	0.2	0.8
9	watermelon	 dark-brown	0.2	0
10	watermelon	 green	0.2	0.8
11	watermelon	 light-brown	0.2	0
12	watermelon	 red	0.2	0
}\datatable

\begin{axis}[
    title=Association Strengths Between Objects and Associated Colors,
    title style={align=center, font=\small},
    axis lines*=left, ymajorgrids,
    width=8cm, height=4cm,
    ymin=0,
    ybar stacked,
    ylabel=Association\\Strength,
    ylabel style={align=center, yshift=-1.5ex, font=\small},
    yticklabel style={font=\small},
    bar width=6pt,
    legend style={at={(0.5,-1), font=\small},
		anchor=north,legend columns=-1},
    xtick=data,
    xticklabels from table={\datatable}{Color},
    xticklabel style={yshift=-5ex,rotate=60,anchor=mid east, font=\small},
    draw group line={Object}{cookie}{cookie}{-4ex}{5pt},
    draw group line={Object}{paper}{paper}{-4ex}{5pt},
    draw group line={Object}{watermelon}{watermelon}{-4ex}{5pt},
    after end axis/.append code={
        \path [anchor=base east, yshift=0.5ex]
            (rel axis cs:0,0) node [yshift=-4ex, font=\small] {Object}
            (rel axis cs:0,0) node [yshift=-8ex, font=\small] {Color};
    }
]

\addplot[ybar, fill=blue!80!white] table [x=X, y=Before] {\datatable}; \addlegendentry{Before Generics}
\addplot[ybar, fill=red!40!white] table [x=X, y=After] {\datatable}; \addlegendentry{Increase After Generics}
\end{axis}
\end{tikzpicture}
\caption{Results of the generic color predicates task, plotting the associations strengths between objects and colors that are associated with them, before and after observing generic input. At first, each object is associated with an arbitrary set of colors that reflect the curriculum. The only associations that increase are correctly the ones observed through generics.}
\label{fig:color-plot}
\end{figure}

\begin{figure*}[t]
\centering
\begin{tikzpicture}
\pgfplotstableread{
X Object	Animal-Similarity	Food-Similarity	People-Similarity	Curriculum
1	wug\\(animal)	0.990	0.196	0	objects-and-kinds
2	vonk\\(food)	0.141	0.981	0	objects-and-kinds
3	snarp\\(people)	0	0	1	objects-and-kinds
4	wug\\(animal)	0.913	0.095	0.336	objects-kinds-and-generics
5	vonk\\(food)	0.112	0.981	0	objects-kinds-and-generics
6	snarp\\(people)	0.345	0	0.928	objects-kinds-and-generics
7	wug\\(animal)	0.991	0.376	0.611	obj-actions-kinds-generics
8	vonk\\(food)	0.401	0.949	0.181	obj-actions-kinds-generics
9	snarp\\(people)	0.611	0.161	0.996	obj-actions-kinds-generics
}\datatable

\begin{axis}[
    title={Similarity of Novel Objects to Each Category across Curricula},
    ybar, axis on top,
    height=3.5cm, width=15cm,
    bar width=0.2cm,
    ymajorgrids, tick align=inside,
    enlarge y limits={value=.1,upper},
    ymin=0, ymax=1,
    axis x line*=bottom,
    axis y line*=left,
    tickwidth=0pt,
    enlarge x limits=true,
    legend style={
        at={(1.1,0.5)},
        anchor=north,
        legend columns=1, font=\small
    },
    ylabel={Similarity},
    ylabel style={yshift=-2ex, font=\small},
    yticklabel style={font=\small},
    xtick=data,
    xticklabels from table={\datatable}{Object},
    xticklabel style={align=center, yshift=-5ex, anchor=north, font=\small},
    draw group line={Curriculum}{objects-and-kinds}{Level 1 (Low)}{-4ex}{7pt},
    draw group line={Curriculum}{objects-kinds-and-generics}{Level 2 (Medium)}{-4ex}{7pt},
    draw group line={Curriculum}{obj-actions-kinds-generics}{Level 3 (High)}{-4ex}{7pt},
    after end axis/.append code={
        \path [anchor=base east, yshift=0.5ex]
            (rel axis cs:0,0) node [yshift=-8ex, font=\small] {Object}
            (rel axis cs:0,0) node [yshift=-4ex, font=\small] {{Complexity}};
    }
]
\addplot[ybar, fill=blue!80!white] table [x=X, y=Animal-Similarity] {\datatable}; \addlegendentry{Animal}
\addplot[ybar, fill=red!40!white] table [x=X, y=Food-Similarity] {\datatable}; \addlegendentry{Food}
\addplot[ybar, fill=orange!5!white] table [x=X, y=People-Similarity] {\datatable}; \addlegendentry{People}
\end{axis}
\end{tikzpicture}
\caption{Results of the category inference task, plotting the similarity of novel objects (e.g \textit{wug}) to animal, food, and people categories across curricula of varying complexity. The correct categories of objects, i.e categories in which they were presented in the generic input, are labeled in parentheses. Regardless of the curriculum complexity and concept category, the learner associates new objects most strongly with the correct category. Incorrect associations increase with curriculum complexity.}
\label{fig:category-plot}
\end{figure*}
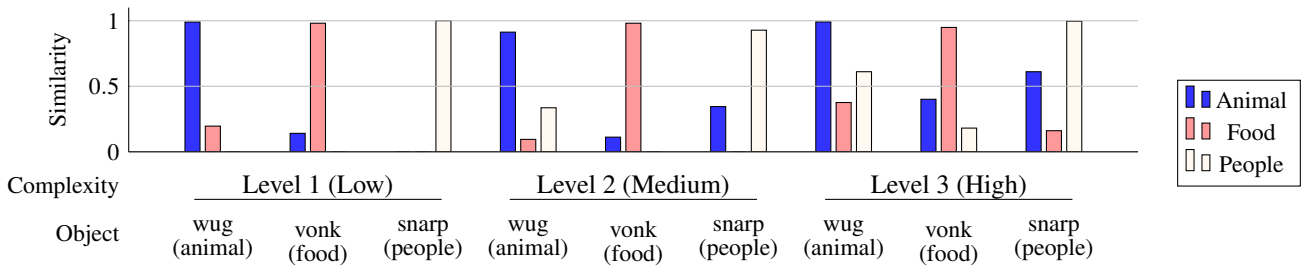

\subsubsection{Results and Discussion}
We measure the association strengths between objects and all associated colors before and after observing generic color predicates. The results are plotted in Figure \ref{fig:color-plot}. Prior to hearing generic input, objects have associations to many arbitrary colors, reflecting the initial non-generic curriculum. Once the generic color predicates are observed, we see that each object has a significantly stronger association with the appropriate color as stated in the generic utterances. Upon observing generics, only the association strengths for the correct colors are updated. These results demonstrate that our model interprets and uses generics as designed, to establish strong associations between an object and a prototypical version of some feature.

\subsection{Task 2: Category Inference}

The category inference task demonstrates how the model forms categories, e.g \textit{animals} and \textit{food}. As visualized in Figure \ref{fig:clustering}, concept categories form naturally in the representational space throughout learning, as similar objects have similar semantic roles in actions; for instance, only animate objects \textit{eat}, and only food objects are \textit{eaten}. In this task, we label these implied categories using generic statements (e.g\textit{``cats are animals," ``dogs are animals"}). Then, we compare the novel concepts placed in these categories with each possible category. We first teach the learner a particular curriculum, then a set of previously unknown objects in known categories (e.g. \textit{``wugs are animals"}), and finally evaluate the similarity of the new item (e.g. \textit{wug}) to each possible category. We repeat this process for different curricula with increasing semantic complexity, by including more complicated content such as \textit{plurals} and \textit{actions}. Table \ref{task-2-tables} shows different learning curricula (top), example utterances (center), and examples with unknown categories (bottom). 
We expected that, while the model should be able to form categories regardless of the curricula, the formed categories would be most distinct in the less complex conditions. 
\begin{table}[h]
\begin{center} 
\caption{Inputs to the learner in the category inference task: (top) learning curricula; (center) examples of situations and utterances; (bottom) inputs presented to the learner. Animal and people categories are treated as separate categories. } 
\label{task-2-tables} 
\vskip 0.12in
\begin{tabular}{lll} 
\hline
 & Learning Curriculum\\
\hline
1 & Objects, categories \\
2 & Objects, categories, generics \\
3 & Objects, colors, actions, plurals, categories, generics \\
\hline
\end{tabular} 
\end{center} 
\end{table}
\vspace*{-\baselineskip}

\begin{table}[h]
\begin{center} 
\begin{tabular}{lll} 
\hline
Curriculum & Examples\\
\hline
Objects & a house; a dog\\
Colors & a red truck; papers are white\\
Actions & a baby drinks milk from a cup\\
Plurals & two balls; many cookies \\
Category Generics & bears are animals \\
Action Generics & cats walk; Moms eat\\
\hline
\end{tabular} 
\end{center} 
\end{table}
\vspace*{-\baselineskip}

\begin{table}[h]
\begin{center} 
\begin{tabular}{lll} 
\hline
New Object & Category & Utterance \\
\hline
wug & animal & wugs are animals \\
vonk & food & vonks are foods \\
snarp & people & snarps are people \\
\hline
\end{tabular} 
\end{center} 
\end{table}
\subsubsection{Results and Discussion}
The results of the category inference task are plotted in Figure \ref{fig:category-plot}. In the easiest curriculum setting containing only objects and categories, every object is most similar to the category in which it was learned, i.e  \textit{wugs} are most similar to \textit{animals}, \textit{vonks} to \textit{foods}, and \textit{snarps} to \textit{people}. Expectedly, we see some similarity between animals and foods, because \textit{chicken} is a member of both categories. The second curriculum condition, which includes objects, categories, and generics, shows that while the correct trend holds, there is some association between animals and people due to generic statements that apply to both categories, such as \textit{sitting}. Finally, in the most complex curriculum condition that includes objects, some actions, plurals, colors, categories, and generics, we see that the associations across categories are increased, but the correct category still has the highest association strength. There is some similarity between animals and foods, and people and animals due to shared concepts such as \textit{chicken} and \textit{eating} respectively. Overall, while associations between object concepts and the incorrect categories increase with curriculum complexity, regardless of the curriculum complexity and category of the novel object, objects are associated most strongly with the correct category in which they were learned through generic statements.

\subsection{Task 3: Joint Category}

The goal of the joint category task is to demonstrate how the ADAM system learns categories across curricula with different contents. Specifically, we create learning curricula for each of the four conditions shown in Table \ref{curriculum-contents-table} and evaluate the model behavior. We use combinations of \textit{chicken}, \textit{beef}, and \textit{cow} objects; \textit{chicken} is used as an example of a lexical item that is shared by two very similar yet distinct concepts (chicken as an animal, and chicken as food) and hence is a member of both the \textit{food} and \textit{animal} categories. While \textit{beef} and \textit{cow} could refer to the same thing at a certain semantic level, they are regarded as disjoint examples of a food and an animal category respectively. We hypothesized that while observing \textit{chicken} would cause some semantic association between food and animal categories, the model should not show any semantic association between food and animal categories when it observes just beef and cow and no chicken.

To execute the task, we run four different versions of the ADAM system, each one trained with one of the four curriculum conditions shown in Table \ref{curriculum-contents-table}. Then, similar to the category inference task, each system is presented a set of previously unknown objects in known categories, e.g “\textit{wugs are animals}”. Finally, we evaluate the similarity of the new object concept to animal and food categories.

\begin{table}[H]
\begin{center} 
\caption{Curricula and contents for the joint category task} 
\label{curriculum-contents-table} 
\vskip 0.12in
\begin{tabular}{ll} 
\hline
  & Test Objects Included in the Learning Curriculum\\
\hline
1 & None (no chicken, beef, or cow) \\
4 & Beef (\textit{food}) and cow (\textit{animal}) \\
3 & Chicken (\textit{food} and \textit{animal})\\
4 & Chicken, beef, and cow \\
\hline
\end{tabular} 
\end{center} 
\end{table}
\begin{figure}[h]
\centering
\begin{tikzpicture}
\pgfplotstableread{
X	Animal-Similarity	Food-Similarity	Curricula-Content
1	0.991965983	0	None
2	0.997029206	0	{Beef\\and Cow}
3	0.993005827	0.477896999	Chicken
4	0.990172229	0.26934181	{Chicken,\\Beef,\\and Cow}
}\datatable

\begin{axis}[
    title=Similarity of the Wug Concept to Animal \\ and Food Categories across Curricula,
    title style={align=center},
    ybar, axis on top,
    height=3.5cm, width=8cm,
    bar width=0.2cm,
    ymajorgrids, tick align=inside,
    enlarge y limits={value=.1,upper},
    ymin=0, ymax=1,
    axis x line*=bottom,
    axis y line*=left,
    tickwidth=0pt,
    enlarge x limits=true,
    legend style={at={(0.5,-0.6), font=\small},
		anchor=north,legend columns=-1},
    ylabel={Similarity},
    ylabel style ={font=\small, yshift=-2ex},
    yticklabel style={font=\small},
    xtick=data,
    xticklabels from table={\datatable}{Curricula-Content},
    xticklabel style={align=center, anchor=north, font=\small, yshift=-1ex},
    after end axis/.append code={
        \path [anchor=base east, yshift=0.5ex]
            (rel axis cs:0,0) node [yshift=-4ex, font=\small] {Curriculum};
    }
]
\addplot[ybar, fill=blue!80!white] table [x=X, y=Animal-Similarity] {\datatable}; \addlegendentry{Similarity to Animals}
\addplot[ybar, fill=red!40!white] table [x=X, y=Food-Similarity] {\datatable}; \addlegendentry{Similarity to Foods}
\end{axis}
\end{tikzpicture}
\caption{Results of the joint categories task. Through generic input, a novel object, \textit{wug} is learned as an \textit{animal}. The plot shows the similarity between the wug concept and the animal and food categories. Wug is more similar to foods when the curriculum includes chicken. The similarity is lower when beef and cow are added and is zero without chicken. }
\label{fig:joint-plot}
\end{figure}
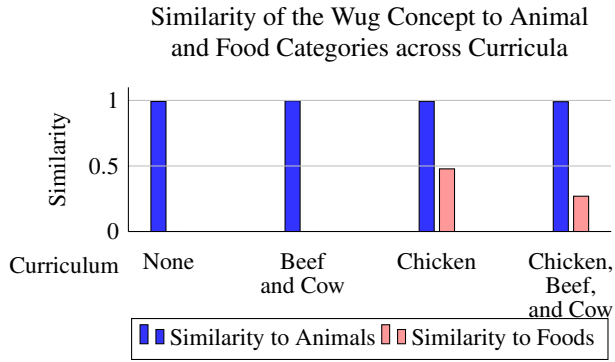

\subsubsection{Results and Discussion}
Figure \ref{fig:joint-plot} plots the results of the joint category task, showing the similarity of the \textit{wug} concept to animal and food categories across curricula with different contents. In the first condition, with a curriculum that does not include \textit{chicken}, \textit{beef}, or \textit{cow}, the similarity to animals is strong, but the similarity level for food category indicates an absence of similarity between animals and foods. Likewise, in the second curriculum condition, when the curriculum includes \textit{beef} and \textit{cow}, but not \textit{chicken}, the learner does not learn any association between food and animal categories. However, when we introduce \textit{chicken} into the curriculum, the association between animals and foods grows. Finally, including \textit{chicken}, \textit{beef}, and \textit{cow} in the curriculum leads to some similarity between foods and animals, but less so than when only \textit{chicken} was included. That these changes match our expectations suggests that ADAM can successfully capture differences in the curricula and the contents in them. While doing so, the system maintains its robustness as indicated by how the similarity to the correct category outscores the similarity to the incorrect category in every condition.

\section{Conclusion \& General Discussion}
We have illustrated that the grounded language acquisition system of ADAM, coupled with the learner module and concept network, can be successfully used for modeling semantics of generic learning. We demonstrated that the model learns from generic language, makes desired semantic associations between learned concepts, and forms semantic categories that reflect the contents of the training curricula. 

The concept network makes it possible to perform operations on learned concepts and on the semantics of generics, enabling generic utterances to establish associations between concepts and consequently form semantic categories. In the future, we plan to further examine the role of association weights in category formation, and how the system performs on different languages. We also hope to explore how we can use the matrix representation of the concept network to integrate ADAM's structured and interpretable meaning representations with distributional models that operate on continuous space, such as neural networks. Moreover, while we did not include the meaning patterns of the learned concepts in the concept network, the flexible nature of the network makes this possible, creating the potential for further semantic analyses that utilize perceptual properties of learned concepts. 

Overall, our system's ability to acquire word meanings and robustness in capturing desired semantic properties across diverse learning curricula shows its promising capacity to model different aspects of language acquisition and to investigate the question of how children discover the expression of generics in their own languages.

\section{Acknowledgements}

Approved for public release; distribution is unlimited. This material is based upon work supported by the Defense Advanced Research Projects Agency (DARPA) under Agreement No. HR00111990060. The  views  and  conclusions  contained herein are those of the authors and should not be interpreted as necessarily representing the official policies or endorsements, either expressed or implied, of DARPA or the U.S. Government.

\bibliographystyle{apacite}

\setlength{\bibleftmargin}{.125in}
\setlength{\bibindent}{-\bibleftmargin}

\bibliography{references}

\end{document}